\documentclass[letterpaper, 10 pt, conference]{ieeeconf}
\IEEEoverridecommandlockouts

\usepackage{graphics}
\usepackage{epsfig}
\usepackage{mathptmx}
\usepackage{times}

\usepackage{color}
\usepackage{subfig}

\usepackage{comment}

\usepackage{epstopdf}
\usepackage{cite}
\usepackage{amsmath,amssymb,amsfonts}
\usepackage{siunitx}  
\usepackage[noend,ruled,vlined]{algorithm2e}
\usepackage{graphicx}
\usepackage{textcomp}
\usepackage{xcolor}
\def\BibTeX{{\rm B\kern-.05em{\sc i\kern-.025em b}\kern-.08em
    T\kern-.1667em\lower.7ex\hbox{E}\kern-.125emX}}

\title{\LARGE \bf
Relative Visual Localization for Unmanned Aerial Systems
}

\author{Steffen Holter, 
Athanasios Tsoukalas,
Nikolaos Evangeliou, Nikolaos Giakoumidis, and
Anthony Tzes
\thanks{The authors are with New York University Abu Dhabi, Electrical and Computer Engineering, P.O. 129188, Abu Dhabi, United Arab Emirates. Corresponding author's
email: \small\tt{anthony.tzes@nyu.edu}}%
}
\begin{document}
\maketitle
\thispagestyle{empty}
\pagestyle{empty}

\begin{abstract}

Cooperative Unmanned Aerial Systems (UASs) in GPS-denied environments demand an
accurate pose-localization system to ensure efficient operation. In this paper we present a novel visual relative localization system capable of monitoring a 360$^o$ Field-of-View (FoV) in the immediate surroundings of the UAS using a spherical camera. Collaborating UASs carry a set of fiducial markers which are detected by the camera-system. The spherical image is partitioned and rectified into a set of square images. An algorithm is proposed to select the number of images that balances the computational load while maintaining a minimum tracking-accuracy level. The developed system tracks UASs in the vicinity of the spherical camera and experimental studies using two UASs are offered to validate the performance of the relative visual localization against that of a motion capture system.

\end{abstract}


\section{Introduction}

The usage of UASs such as drones has become increasingly commonplace due their ability to accomplish basic tasks in fields ranging from transport to surveillance \cite{semsch2009autonomous}. Yet, to tackle complex assignments such as Search and Rescue (SaR) missions \cite{nourbakhsh2005human} and intruder detection \cite{howard2006experiments} requires the utilization of multiple distributed smaller drones in collaboration \cite{wang2017taking}. While using collaborative drone swarms is a cost-effective and reliable option, there are also critical design challenges associated with creating efficient multi-UAS systems. 

The most immediate issue being the need to effectively convey the real-time relative positions and orientations of the team members. This creates the foundation for the implementation of multi-agent coverage control and ensures timely collaboration between the distributed drones. In particular, the relative pose information needs to be reported with minimal latency \cite{mahdoui2016multi}. While communication-based models can cope with these problems, not all UASs are equipped with such capabilities or may be operating in conditions where they are impaired \cite{balamurugan2016survey}. Vision based detection represents an effective alternative as optical sensors and cameras have become increasingly more prevalent. Similarly, in collaborative efforts the focus has shifted from endogenous pose estimation techniques such as GPS and Inertial Measurement Units (IMU) to exogenous ones including LIDAR and Receive Signal Strength Indicator (RSSI) \cite{balamurugan2016survey,martinez2011board,deilamsalehy2016sensor}. 

In this paper, we present a novel visual pose estimation system that makes use of a spherical camera. By optimally splitting and rectifying the recorded spherical image into equal sized square partitions the entire surrounding space can be monitored simultaneously in real-time. Similarly, any collaborating UASs in the nearby vicinity will be detected within a specified distance. The detection and pose estimation of the target drone is performed using a rhombicuboctahedron fiducial marker set \cite{tsoukalas2018relative} which is mounted on the target collaborating drones. 

Due to the high computational demands associated with searching the entire surrounding environment an optimized search algorithm is used to parse the sphere partitions based on the last seen location. To further improve performance, shifting the image rectification to the GPU allows for the visual processing to occur in parallel. These modifications ensure that the space can be monitored in real-time. 

To exemplify the effectiveness and test the limitations of the proposed localization system, an experimental study is conducted. By equipping one drone with the spherical camera and the other with the fiducial marker the ability to identify the target drone can be determined. A high accuracy motion capture system is used to verify the visual pose measurements made by the spherical camera. Performing an experiment in such a controlled environment demonstrates the practical application of the system.
\section{Related Work}
Work in visual UAS pose estimation and detection can be broadly classified into two main groups: marker based and non-marker based. Thus far, the majority of UASs rely on on-board artificial markers for localization \cite{jin2019drone}. These systems might include a fiducial marker identification scheme that utilizes a special arrangement of squares inside a rectangle in order to identify and distinguish markers \cite{garrido2016generation, garrido2014automatic}. Common examples of this are ArUco \cite{munoz2012aruco}, AprilTags \cite{olson2011apriltag,wang2016apriltag}, and Alvar \cite{alvar2016library}. Another marker example used mounted LEDs that are detected using infrared cameras \cite{su2017uav}. All these frameworks demonstrate flexibility and simplicity in application while yielding notable results. However, making use of predetermined markers requires the targets to be identified and tagged prior to detection. In other words, their feasibility is largely limited to collaborative efforts. 

Non-marker based systems aim to perform similar pose estimates without any fixed reference points. On-board visual pose estimation for UASs usually are either odometry-based approaches that use optical flow, or simultaneous localization and mapping (SLAM) \cite{harmat2015multi,lu2018survey}. In addition, recent work in this field has largely focused on integrating deep leaning based solutions \cite{brachmann2018learning,mahendran20173d,unlu2019deep}. These frameworks can also scale to non-collaborative systems, but are much more situation dependent and difficult to implement.

\section{Problem Definition}
In any collaborative drone swarm system, knowledge of the relative pose of neighbouring UASs is imperative. However, recovering this information with sufficient accuracy can be difficult especially in close proximity \cite{woods2015dynamic}. Not only are classical methods like GPS-based systems inaccurate they also might not always be available. However, in GPS-denied environments visual sensors present a viable and cost-effective alternative for acquiring this information. By determining the pose locally the need for direct transmission of location information between the UASs in the system is also eliminated. However, such an approach has the obvious limitation that neighboring UASs have to be within line-of sight of each other to estimate the pose.

Furthermore, while single 2D cameras can successfully perform localization through the use of markers or deep learning based models (e.g. convolutional neural networks), their FoV is limited. Increasing sensing capabilities by adding additional cameras can be both costly and computationally more demanding. Thus, without adding a supplemental system to notify of the proximity of collaborator drones, visual pose detection is difficult. 

The primary contribution of this paper is presenting a combined system that is able to monitor the entire immediate surroundings of the UAS as well as perform drone localization. This is achieved through the effective integration of a spherical camera~\cite{aghayari2017geometric} with a near $360^{\circ}$ FoV. Minimal on-board computational capabilities are required to provide real-time operation. Additionally, the proposed system extends beyond UAS identification and can theoretically be implemented in any collaborative framework. 

Furthermore, the system adopts an algorithmic approach to minimize computational requirements and optimally search the surroundings by monitoring the previous position of the UAS. This allows the system to achieve a rapid pose estimation rate of nearly 30 fps with an Intel i7-system. 
\section{Proposed System}
\begin{figure}[h]
  \centering
  \includegraphics[width=0.85\linewidth]{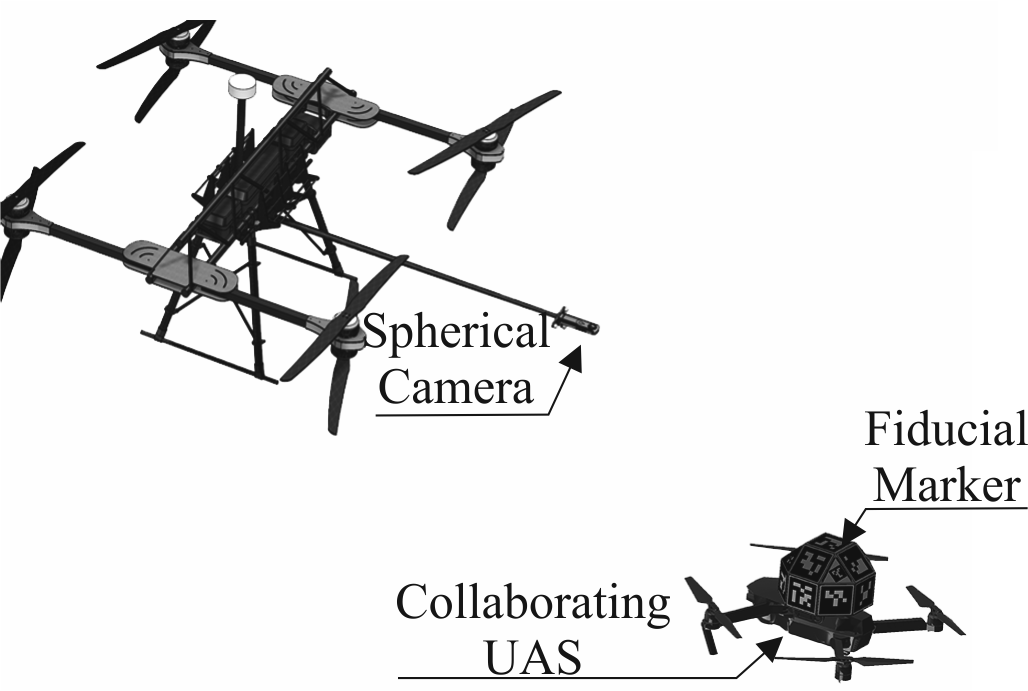}
  \caption{Visual Relative Localization System}
  \label{fig:exp-setup}
\end{figure}

The suggested system assumes that all collaborating drones carry a set of fiducial markers and a spherical camera for detecting their neighbors as shown in Fig.~\ref{fig:exp-setup}.

\begin{figure}[h]
  \centering
  \includegraphics[width=0.9\linewidth]{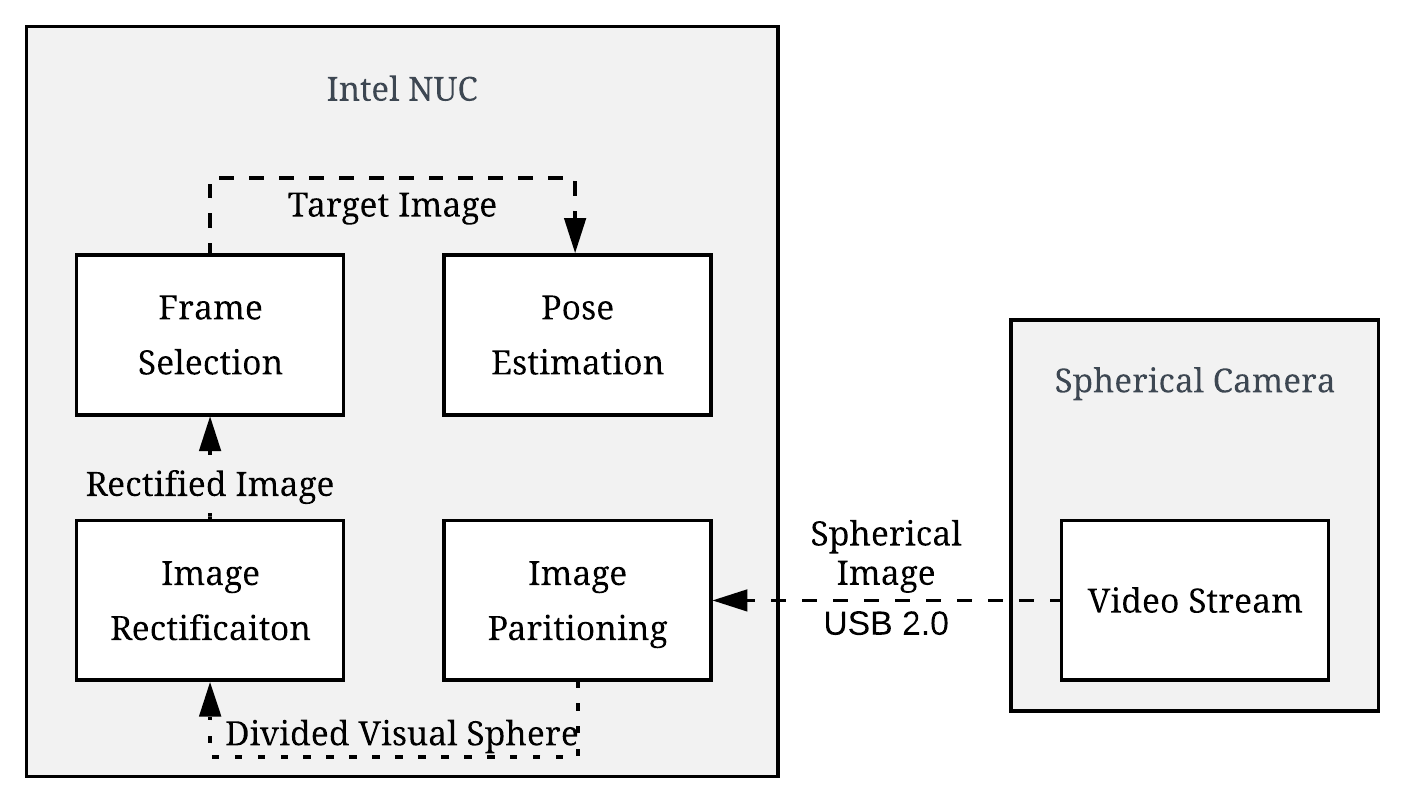}
  \caption{UAS-relative localization software composition}
  \label{fig:soft-comp}
\end{figure}

 The developed system can be divided into three main components: partitioning the spherical image, image rectification and pose estimation. The resulting combination is a novel visual UAS-localization system summarized in Fig.~\ref{fig:soft-comp}.

\subsection{Spherical Image Partitioning}

In order to be able to accommodate standard UAS localisation algorithms the spherical image needs to be rectified to a classic distortionless rectilinear format, comparable to a pinhole camera \cite{young1971pinhole}. By default the spherical camera records images in a "spherical format" which is comprised of two wide-angle frames stitched together to form a virtual sphere \cite{aghayari2017geometric}. Hence, it is not possible to make a direct conversion to our desired standard format. To retain all of the visual information, it is crucial to split this spherical image into smaller partitions which can then be rectified to simulate a pinhole camera. Essentially, the challenge becomes tessellating the surface of a spherical image into equal sized square segments. Most established methods to approximate such a split make use of 2D projections to mimic the spherical shape, yielding largely asymmetrical non-square cells \cite{rossow1984selection,gringorten1992division,beckers2012general,beckers2011universal}. It should be noted that there is no optimal method to tessellate the sphere's surface into equal area square partitions \cite{saff1997distributing}. 

\begin{figure}[h]
  \centering
  \includegraphics[width=0.5\linewidth]{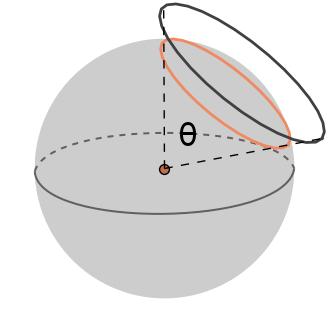}
  \caption{Spherical projection of a tangent circle with angle $\theta$}
  \label{simple-sphere}
\end{figure}

\begin{figure*}[h]
  \centering
  \includegraphics[width=0.77\linewidth]{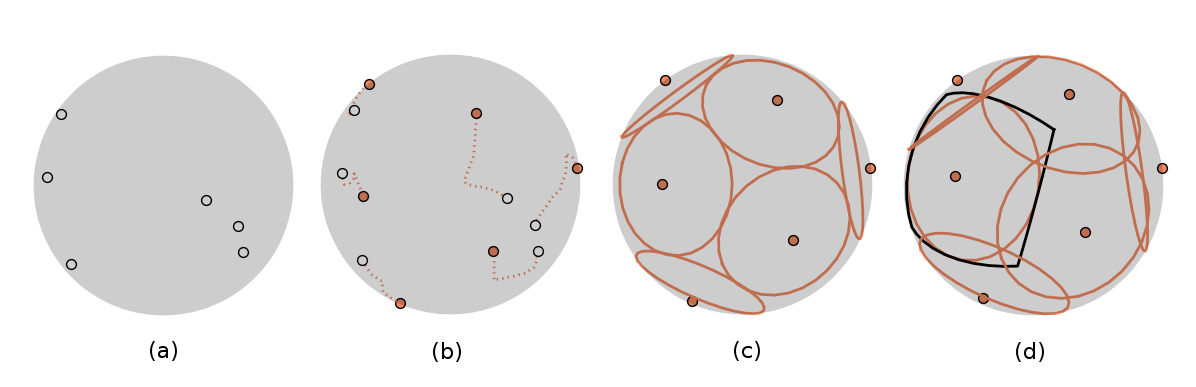}
  \caption{Sphere partitioning algorithm for $N=12$ \textcircled{a} point initialization, \textcircled{b} trajectory for equal distribution, \textcircled {c} optimal packing of circles, \textcircled{d} full coverage with circles,including conversion to square projection.}
  \label{fig:sphere-comp}
\end{figure*}

Therefore, rather than attempting the tessellation of a sphere's surface with squares, orientation-independent circles will be used as an initial approximation. This is due to the arbitrary conversion that can then be made later from circular frames to square ones. This acts as an extension of the Tammes problem which is defined as finding the arrangement of $N$ points on a sphere that maximizes the minimum distance between any two points \cite{musin2015tammes}. For our system, however, it is important to also avoid any gaps in the visual detection. In other words, we aim to distribute a certain number of circles in a way that ensures total coverage while minimizing overlap. As such, let $N$ be the number of circles used and $\theta$ the polar (spherical) angle that reflects the minimum diameter of the circles needed to completely cover the sphere as shown in Fig.~\ref{simple-sphere}.

To calculate the "diameter" angle $\theta$ given a fixed $N$, our system relies on a randomized algorithm complemented with fine-tuning through numerical computational logic. The circle centers are initialized randomly across the surface of the sphere as in Fig.~\ref{fig:sphere-comp}\textcircled{a}. Through an iterative process, the two closest centers are separated by a constant factor until convergence as shown in Fig.~\ref{fig:sphere-comp}\textcircled{b}. This results in a near-optimal uniform distribution of points on the surface of the sphere as in Fig.~\ref{fig:sphere-comp}\textcircled{c}. 
The angular size of the circles is increased until every single point in the sphere's surface falls within the range of least one circle center (Fig.~\ref{fig:sphere-comp}\textcircled{d}). 

With the aim of implementing the spherical image rectification in real-time it is necessary to identify the $N$ number that results in the best accuracy result and lowest computational overhead. The run-time is directly correlated to the the number of pixels that need to be processed or, in other words, the amount of surplus overlap resulting from each configuration. However, when constructing the optimization function it is also important to consider qualitative factors such as image distortion. 

Firstly, to find the total number of single pixels processed ($p$) in the rectification process, we compute the number of pixels enclosed by each circle on the sphere, defined as
\begin{equation}
p = N \times \frac{(360\si{\degree})^2}{\theta^2} \times (h \times w)~,
\end{equation}
where the pixel count ($h \times w$) is that of the original input spherical image.

\begin{figure}[h]
  \centering
  \includegraphics[width=0.85\linewidth]{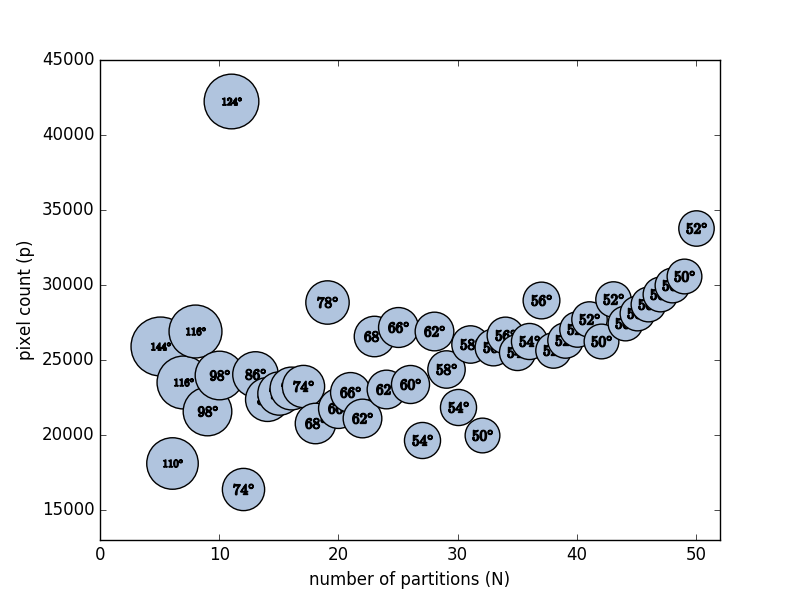}
  \caption{Pixel count and $\theta$ for different partitioning schemes}
  \label{volume-comp}
\end{figure}

Minimizing this quantity would guarantee the computational optimum if the pixel processing was performed sequentially. However, since the image rectification is performed on a GPU the processes will be carried out in parallel. This suggests that increasing the pixel count in a single partition will result in a smaller than expected increase in computational demands. However, since each individual circle is, in fact, processed sequentially this same benefit does not extend to increasing the number of partitions ($N$). Thus, this separate $N$ dependent error factor needs to be accommodated in the optimization function. Furthermore, it is clear that larger partitions exhibit more distortion ($d$) after rectification thus a small $N$ would result in lower quality images. To account for these three main factors the following optimization function criterion is used to compute $N$
\begin{eqnarray}
\min &&  \left [\alpha_1 N + \alpha_2 p + \alpha_3 d\right],~\mbox{where}\nonumber\\
\sum_{i=1}^3 \alpha_i =1,&&\alpha_i >0,i=1,2,3,~\mbox{and}~
\alpha_1 \gg \alpha_2~. \nonumber
\end{eqnarray}
The relationship between $N$ and the pixel count $p$ is depicted in Fig.~\ref{volume-comp}. The size of each point in the graph indicates the $\theta$-angle. The overall trend is constant with a few noticeable outliers. The lowest pixel counts occur at $N=12$ and $N=6$ suggesting minimal overlap and efficient packing of circles on the sphere. While our optimization function leans towards lower $N$ values it is important to also account for the qualitative distortion variable which is significant for $N=6$. Thus, the optimal solution for the system occurs at $N=12$. Fig.~\ref{fig:sphere-comp}\textcircled{d} shows the resulting distribution of circles on the sphere for $N=12$.

\subsection{Image Rectification}

\begin{figure}[h]
  \centering
  \includegraphics[width=0.85\linewidth]{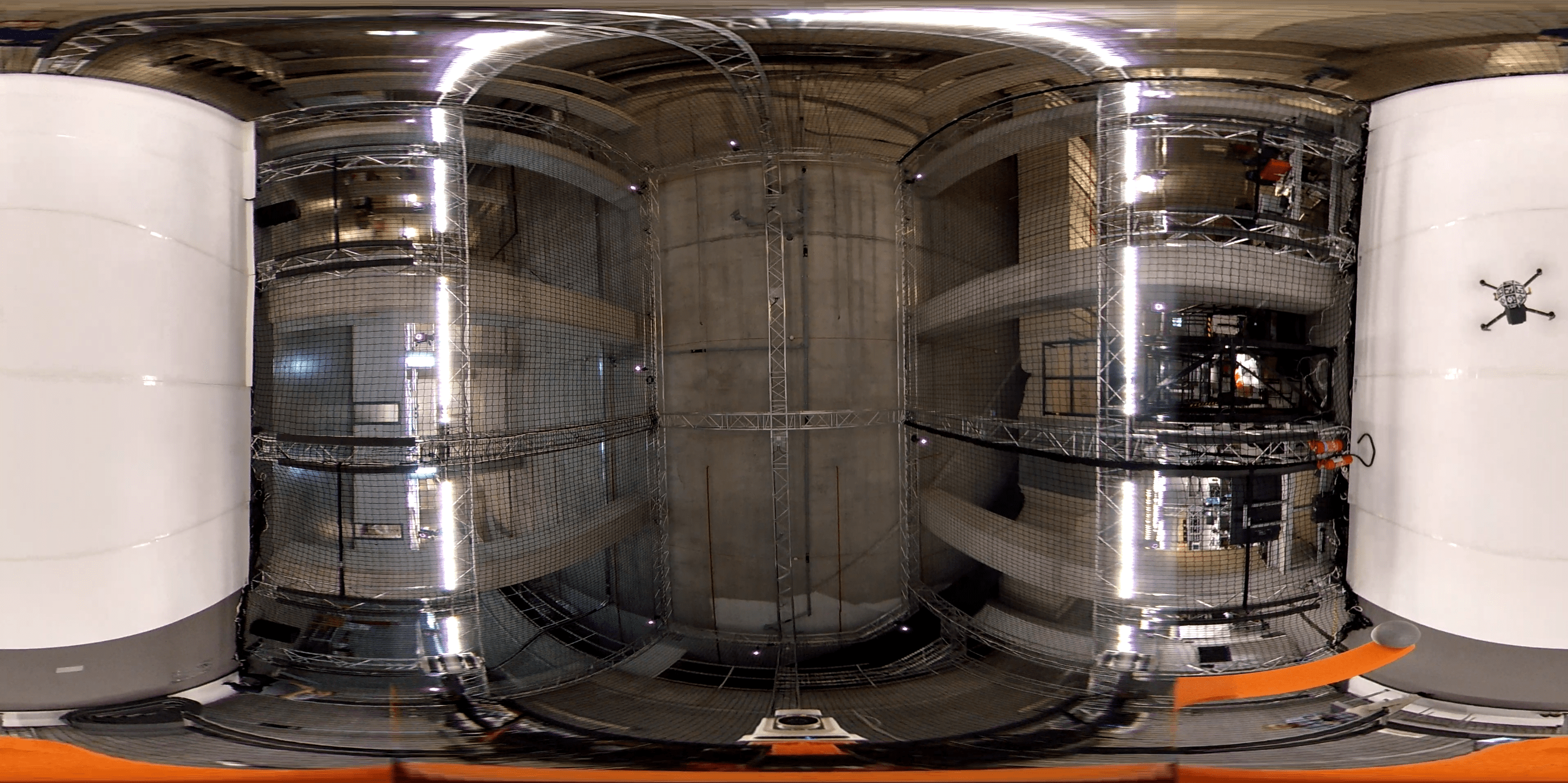}
  \caption{Spherical flat image}
  \label{fig:orig sphere}
\end{figure}

The acquired flat spherical image, shown in a typical frame in Fig.~\ref{fig:orig sphere}, is first translated to the geographic coordinate system by applying inverse stereographic projection \cite{saalfeld1999delaunay,coxeter1961introduction}. Initially, the latitude ($\phi$) and longitude ($\lambda$) for each pixel in the image is found, resulting in a spherical mapping that can be navigated using standard rotation matrices.
\begin{eqnarray}
\rho = \sqrt{x^2 \times y^2} &,& 
c = 2 \arctan{\frac{\rho}{2R}}\nonumber\\
\lambda = \arctan{{\frac{x \sin{c}}{ \rho \cos{c}}}} &,&
\phi = \arcsin{\frac{y \sin{c}}{\rho}}. \nonumber
\end{eqnarray}

The size of each of the individual partitions is found by making the arbitrary conversion from circle to square as can be seen also in Fig.~\ref{fig:sphere-comp}\textcircled{d}. The "diameter" angle of the circle is taken as the pan ($\theta$) and tilt ($\phi$) angles of the square partition. Applying a zoom factor selects the pixels that fall into this specified scope. To isolate the desired part of the sphere, we use the optimal circle centers calculated through the partitioning algorithm. Applying simple rotation matrices centers the partition at the target section of the pixel sphere resulting in the rectified image. This process is repeated for each partition.
As mentioned previously, this method has a clear limitation with regards to the pan and tilt angles. At large angle values, in order to accommodate the increased FoV, the image starts to display signs of distortion as the desired rectilinear property is violated. Essentially, it is gradually reverting back to a wide-angle image. This clearly exemplifies why the entire image cannot be rectified directly through a single conversion.

\subsection{Pose Estimation}

To perform accurate pose estimation in the immediate surroundings, a marker based system is preferred. Since our framework caters to groups of collaborating drones, it is reasonable to make use of fiducial markers. Tsoukalas et al. proposed a fiducial marker system that uses markers from the ArUco library oriented in a rhombicuboctahedron formation~\cite{tsoukalas2018relative}. The pose detection system was determined to have an effective operating range of up to 2.2 meters with the default marker size $(8 \times 8$cm) parameters. This is deemed sufficient in fulfilling the requirement of having effective functionality in the UAS's immediate surroundings. The use of the rhombicuboctahedron marker configuration offers robustness in occlusion and lighting conditions which are beneficial for close range detection. 

For the localization, each sphere partition isolated from the camera feed is checked for ArUco markers. For every identified individual square marker its pose is inferred. Based on all the markers captured the global position and orientation of the truncated rhombicuboctahedron is computed. The more single markers that are captured the more accurate the overall pose estimation. 


As stated previously, parsing through all of the separate partitions is computationally demanding. Using a greedy search that loops through all of the frames results in high latency. Therefore, to allow for real-time detection an optimized search algorithm is integrated into the pose estimation process.

The general framework of the algorithm is summarized below. When initialized, the system must parse all of the frames arbitrarily due to the lack of previous information. However, once the first collaborating UAS is detected, each subsequent search is performed by accounting for that measurement. The sphere partition where the last detection occurred is searched first. If no markers are identified, then the neighbouring frames are searched, after which, the search continues based on closest distance to the previous detection frame.
\begin{algorithm}[h]
\SetAlgoLined
\KwResult{Identify partition with marker}
    $n \leftarrow None$ \\
    \While{video\_feed}{
    $all\_parts \leftarrow partition\_sphere()$ \\
        \uIf{n \textbf{is} None}{
            \For{ $p \in all\_parts$}{
                $det \leftarrow detect\_marker(p)$ \\
                \If{det.found()}{
                    $n \leftarrow det.index()$ 
                }
            }
        }
        \uElse{
        $det \leftarrow detect\_marker(all\_parts[n])$ \\
            \uIf{det.found()}{
                $n \leftarrow det.index()$ 
                }
            \uElse{
            $sorted\_parts \leftarrow sort\_by\_dist()$ \\
                \For{ $p \in sorted\_parts$}{
                    $det \leftarrow detect\_marker(p)$ \\
                    \If{det.found()}{
                        $n \leftarrow det.index()$ 
                    }
                }
            }
        }
    }
 \caption{Optimized Sphere Partition Search}
\end{algorithm}
It should be noted that the algorithm is content with finding the first set of markers it finds. Due to the inherent overlap that comes with the inability to perfectly cover the sphere, multiple image partitions might contain markers. While the first set identified through the search might turn out to be inferior in number to the other sets in the remaining partitions, it is inefficient to keep searching further. To absolutely guarantee the best set would again require parsing every single individual partition for each frame.

\begin{figure*}[h]
  \centering
  \includegraphics[width=0.85\linewidth]{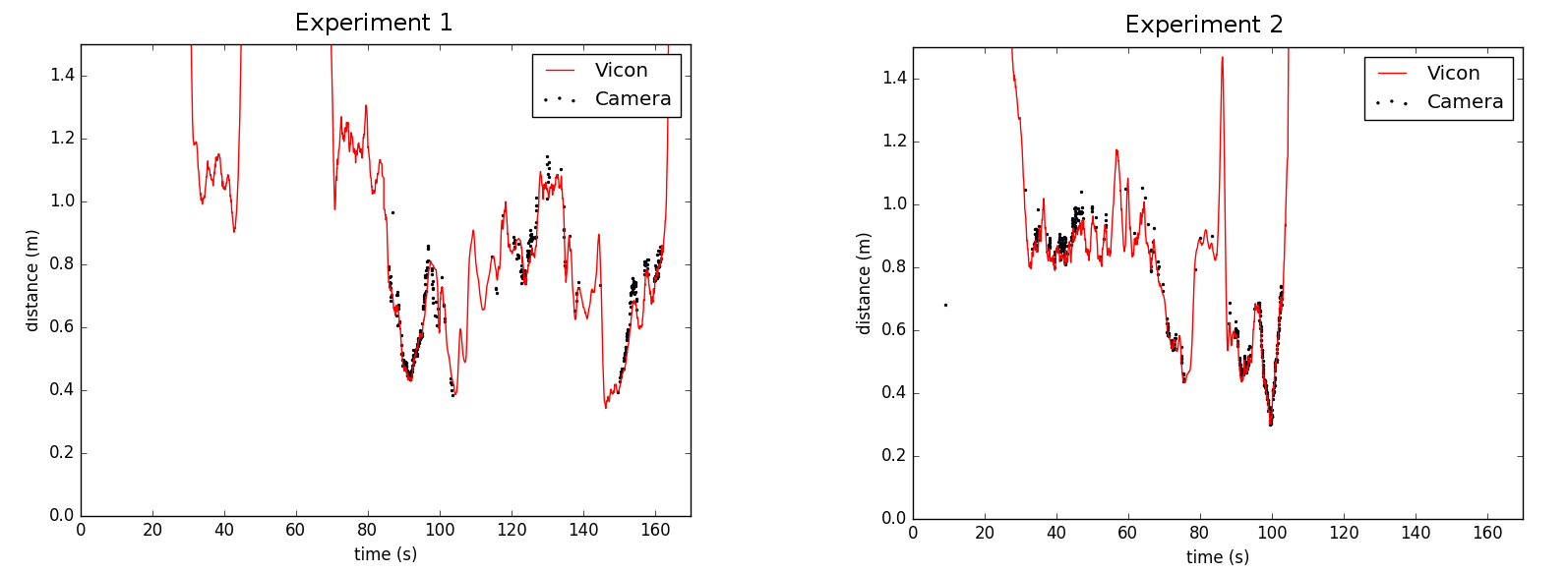}
  \caption{Motion capture vs visual distance measurements for sample experiments}
  \label{fig:exp-results}
\end{figure*}
\section{Experimental Studies}

\begin{figure}[h]
  \centering
  \includegraphics[width=0.85\linewidth]{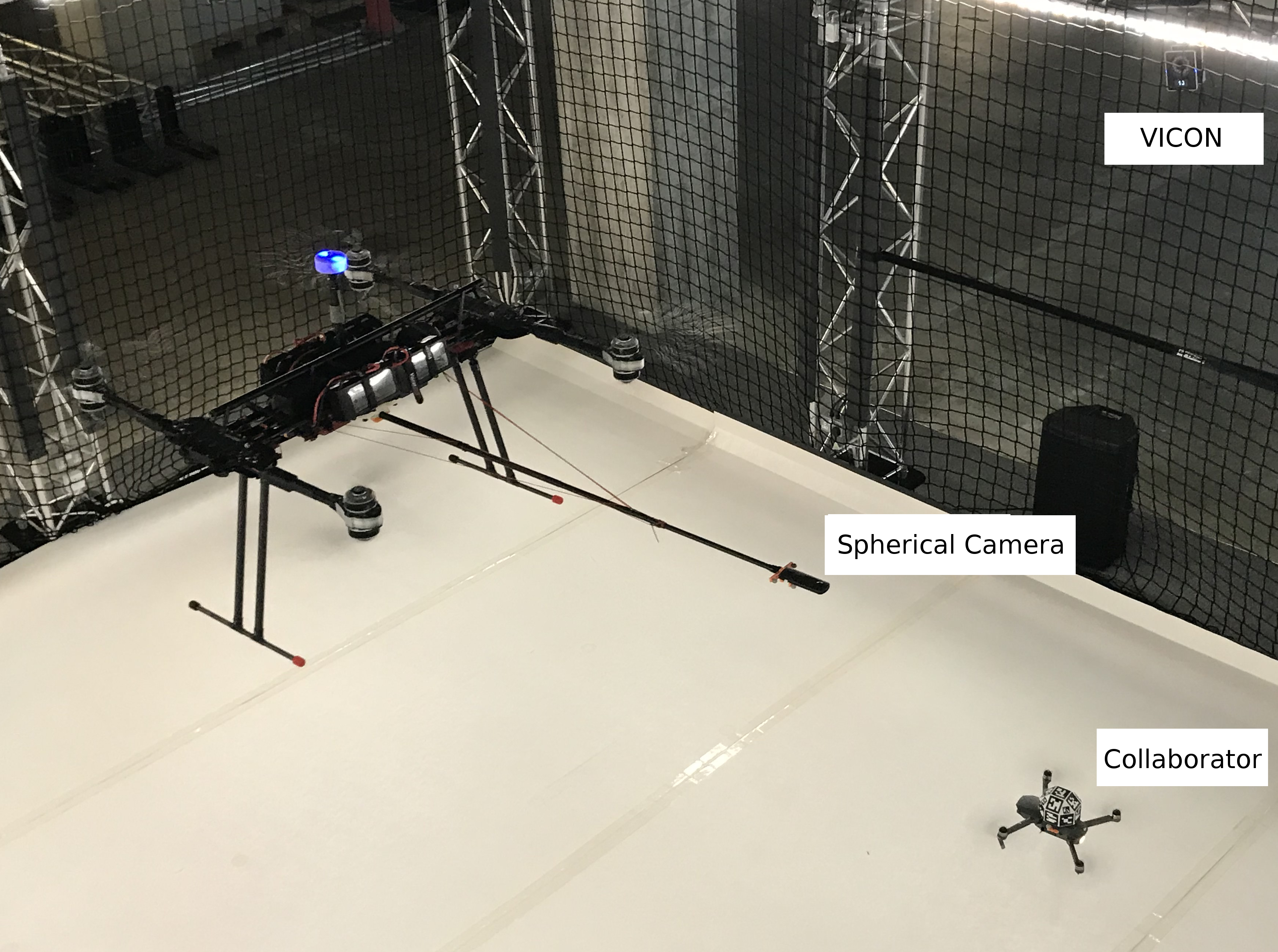}
  \caption{Experimental setup for relative localization}
  \label{fig:exp-live}
\end{figure}

To test the effectiveness of the proposed system an experimental analysis is performed using an octarotor UAS that is detecting the relative pose of a collaborator UAS mounted with fiducial markers, as shown in Fig.~\ref{fig:exp-live}. A direct comparison is made between the pose estimate from the visual spherical camera system and a high accuracy motion capture system.

\subsection{Hardware specifications}

For this experimental study, a Vulcan D8 UAS is used equipped with an i7-Intel NUC system and a basic GPU to facilitate parallel computing. Furthermore, a Ricoh Theta V spherical camera is mounted in front of the drone attached to a protruding 2.7m stick to facilitate efficient indoor testing. On the other hand, a smaller version of the previously described rhombicuboctahedron is attached to a DJI-Mavic drone with $5 \times 5$cm markers. 
Both of the UASs were flown in our Kinesis CTPindoor test facility (15m× 5m × 8m (W×L×H))

The location of the UASs was measured throughout the experimental process with a motion capturing system comprising of 24 Vicon Vintage cameras at a rate of 120Hz. The system achieves sub-millimeter accuracy using reflective markers, which were placed to monitor the pose of the spherical camera and the pose of ArUco marker. 

\subsection{Visual pose detection}

The quadrotor drone was flown in a randomized trajectory near the vicinity of the octarotor one. 
To demonstrate the overall performance of the system we aimed to evaluate the three key components, namely, the sphere partitioning, the partition search and location measurements, separately. These sections are deemed to represent the computationally critical points in the system's workflow and thus their optimization ensures success of the entire framework.

\begin{figure}[h]
  \centering
  \includegraphics[width=0.85\linewidth]{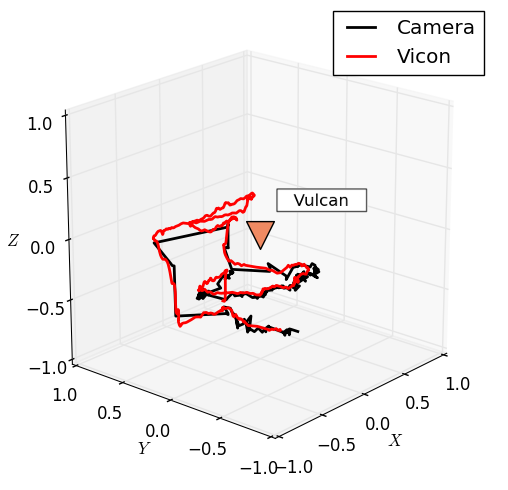}
  \caption{3D flight path for the DJI Mavic relative to Vulcan D8 UAV}
  \label{fig:3d measurements}
\end{figure}

Fig.~\ref{fig:exp-results} shows the distance measurements to the target collaborating drone from the spherical camera and the Vicon motion capture system for two experimental cases. The localization results clearly indicate that the drone is being detected with a very high accuracy within a region of approximately 1.5m. The shortened detection range is due to the reduced size of the marker. However, when detected the absolute localisation error is an average of only 3.8cm for experiment (1) and 2.2cm for experiment (2). Closer inspection reveals periods where the drone is within the expected detection range, however, the camera is unable to recognize the marker. A cross-analysis of the individual x,y,z coordinate measurements shows that these instances occur mostly when the drone's flight path goes above the camera meaning the marker is blocked from the camera's FoV by the body of the UAS. An example of the recorded flight path for the DJI MAviq drone maneuvering around the hovering octarotor can be seen in Fig.~\ref{fig:3d measurements}. The example isolates a 40 second interval (85-145s) from experiment 1 to ensure clarity and minimize the crossing of flight paths. Comparing the Vicon and camera system measurements in this view further confirms successful detection of the target drone.

\begin{figure}[h]
  \centering
  \includegraphics[width=0.85\linewidth]{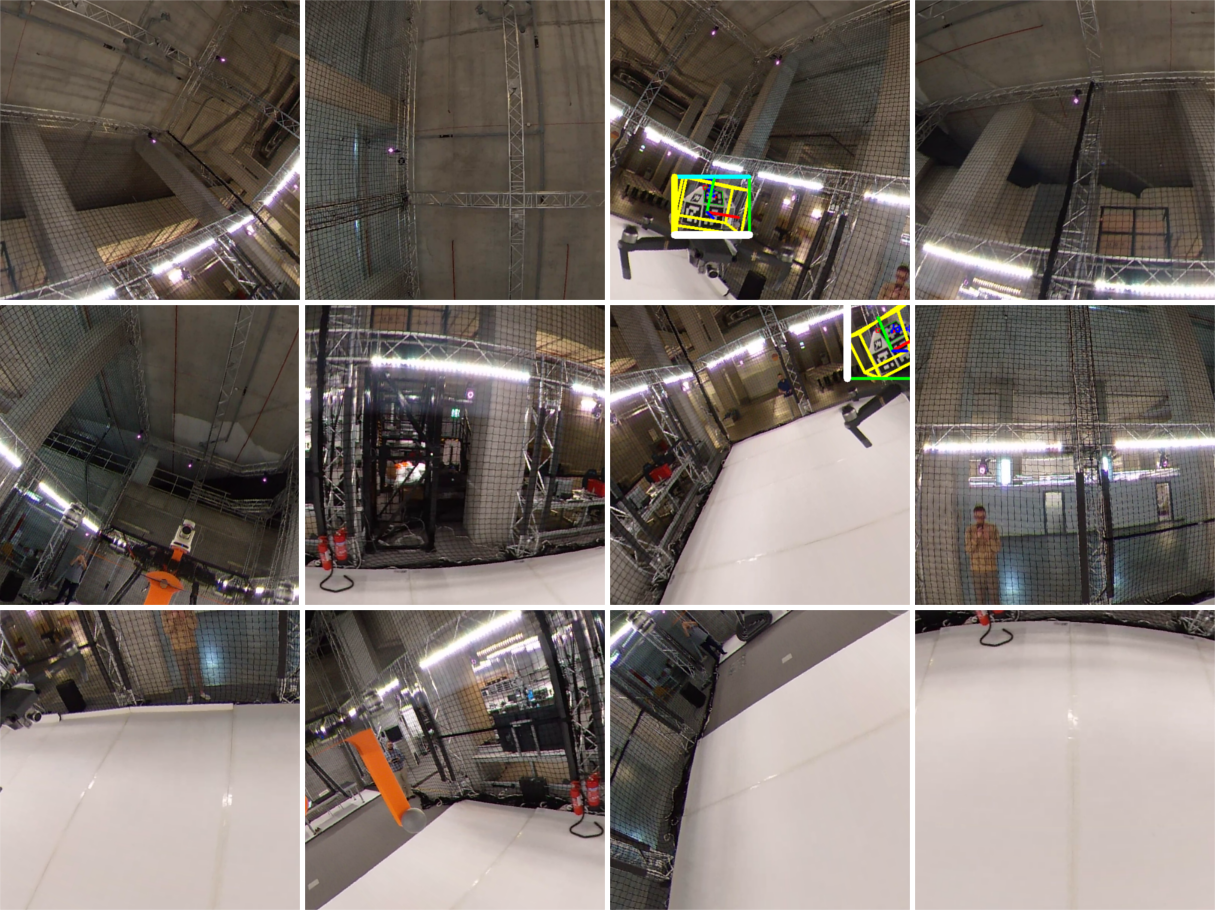}
  \caption{$N=12$ rectified partitions for a single video frame}
  \label{fig:frames}
\end{figure}

To evaluate the proposed optimal sphere partitioning solution, the same raw video feed from experiment 1 is processed using two alternative $N$ values. $N=6$ and $N=24$ were chosen as representatives of small and large frame partitioning options respectively. A summarizing analysis can be seen in Table \ref{sphere-comp}. Comparing the average Frames-per-Second (FpS) suggests a considerable decrease as $N$ is increased. This was expected and confirms the influence of parallel processing. However, while at $N=6$ the speed is optimal there is a steep decrease in performance. Only 281 localisation measurements were made over the course of the experiment and the average error with respect to the Vicon motion capture system was 9.9cm. This is due to the significant distortion that results from rectifying large partitions. Our chosen $N=12$ configuration by far out performs both extremes when it comes to accuracy and frames captured. At $N=24$ the FpS is very low and the prediction accuracy also falls below our optimal solution. This is mainly due to many markers falling on the borders of the smaller partitions and thus resulting in false predictions. The ideal configuration for $N=12$ is depicted in Fig.~\ref{fig:frames} as each of the rectified partitions for a single video frame from the experiment are shown. This particular example also demonstrates that even for our optimized split there is still inherent overlap in the coverage, as the drone is identified in two separate sections.

\begin{table}[h!]
\renewcommand{\arraystretch}{1.3}
\caption{Sphere partition comparison}
\label{sphere-comp}
\centering
\begin{tabular}{c c c c}
\hline
& \textbf{N=6} & \textbf{N=12} & \textbf{N=24}\\

\hline
Angle ($\theta$) & 110 & 74 & 62 \\
FpS & 6 & 4.5 & 2.8 \\
Error (cm) & 9.9 & 3.8 & 6.6 \\
Frames Captured & 281 & 395 & 299 \\
\hline
\end{tabular}
\end{table}

Lastly, the computational benefit derived from the optimized partition search is exemplified through a comparison with a standard greedy parsing algorithm, that arbitrarily loops through the partitions. Table~\ref{algo_comp} compares the number of localization measurements measured using the two algorithms. This measure is used since, as per the design, the benefit of the optimisation is only evident during and after periods of detection. Additionally, to contextualize the results the total number of image frames in the video feed that can theoretically yield a pose estimate was found through a frame-by-frame analysis. An increase of 25\% was achieved in the number of localization measurements through the implementation of the algorithm. However, as expected even using the improved method 60\% of the total frames in which a UAS could possibly be detected are lost, due to the basic computational demands of the framework.

\begin{table}[h!]
\renewcommand{\arraystretch}{1.3}
\caption{Search Algorithm Comparison}
\label{algo_comp}
\centering
\begin{tabular}{c c c}
\hline
 & \bfseries Greedy Search & \bfseries Optimized Search\\
\hline
Numbers Recorded & 147 & 395 \\
Total Available & 986 & 986 \\
Percentage Recorded & 15\% & 40\% \\
\hline
\end{tabular}
\end{table}




\section{Conclusion}

In this paper we introduce a novel method for 360 degree UAS localisation using fiducial markers. Through the optimized partitioning of the spherical field-of-view the UAS can detect any collaborating drones in near-proximity. Through the integration of a targeted search the computational overhead is minimized and the system is applied in real time. Experimental studies showed the relative localization to be accurate to within approximately 4cm.


\bibliographystyle{ieeetr}
\bibliography{main}



\end{document}